\documentclass[conference]{IEEEtran}
\IEEEoverridecommandlockouts
\usepackage{cite}
\usepackage{booktabs}
\usepackage{algorithm}
\usepackage{algpseudocode}
\usepackage{amsmath,amssymb,amsfonts}
\usepackage{graphicx}
\usepackage{textcomp}
\usepackage{xcolor}
\usepackage{array}
\usepackage{fancyhdr}
\usepackage{subcaption}
\usepackage{lipsum}

\usepackage{booktabs}
\usepackage{multirow}

\def\BibTeX{{\rm B\kern-.05em{\sc i\kern-.025em b}\kern-.08em
    T\kern-.1667em\lower.7ex\hbox{E}\kern-.125emX}}
\begin{document}

\title{Attention-Based Offline Reinforcement Learning and Clustering for Interpretable Sepsis Treatment}

\author{\IEEEauthorblockN{Punit Kumar, Vaibhav Saran, Divyesh Patel, Nitin Kulkarni, and Alina Vereshchaka}
\IEEEauthorblockA{Department of Computer Science and Engineering \\
University at Buffalo \\
Buffalo, New York, USA}
\IEEEauthorblockA{\{punitkum, vsaran, dpatel45, nitinvis, avereshc\}@buffalo.edu}
}

\maketitle

\thispagestyle{fancy}
\fancyhf{} 
\renewcommand{\headrulewidth}{0pt} 
\fancyfoot[c]{%
  \parbox{\textwidth}{%
  \centering \scriptsize
  \copyright~2025 IEEE. Personal use of this material is permitted. Permission from IEEE must be obtained for all other uses, in any current or future media, including reprinting/republishing this material for advertising or promotional purposes, creating new collective works, for resale or redistribution to servers or lists, or reuse of any copyrighted component of this work in other works.
  }
}

\begin{abstract}

Sepsis remains one of the leading causes of mortality in intensive care units, where timely and accurate treatment decisions can significantly impact patient outcomes. In this work, we propose an interpretable decision support framework. Our system integrates four core components: (1) a clustering-based stratification module that categorizes patients into low, intermediate, and high-risk groups upon ICU admission, using clustering with statistical validation; (2) a synthetic data augmentation pipeline leveraging variational autoencoders (VAE) and diffusion models to enrich underrepresented trajectories such as fluid or vasopressor administration; (3) an offline reinforcement learning (RL) agent trained using Advantage Weighted Regression (AWR) with a lightweight attention encoder and supported by an ensemble models for conservative, safety-aware treatment recommendations; and (4) a rationale generation module powered by a multi-modal large language model (LLM), which produces natural-language justifications grounded in clinical context and retrieved expert knowledge. Evaluated on the MIMIC-III and eICU datasets, our approach achieves high treatment accuracy while providing clinicians with interpretable and robust policy recommendations. 
\end{abstract}


\section{Introduction}
Sepsis is a life-threatening medical emergency characterized by a dysregulated host response to infection, leading to acute organ dysfunction and high risk~\cite{Singer2016}. In-hospital risk rates range from approximately $10-30\%$, and can exceed $40\%$ in cases of septic shock, underscoring the profound lethality of this condition~\cite{Rhee2017}. In the United States, sepsis affects over $1.7$ million adults annually and accounts for more than a third of hospital deaths, imposing a substantial human and economic burden~\cite{Torres2023}. The rapid progression of sepsis and its heterogeneous presentation across diverse patient populations underscore the urgent need for early detection, precise diagnosis, and timely intervention to improve patient outcomes~\cite{Evans2021}.

Recent advances in large-scale critical care databases such as MIMIC-III~\cite{johnson2016mimic} and eICU~\cite{Pollard2018}, combined with machine learning (ML) and reinforcement learning (RL) methods, offer promising tools for personalized treatment planning. While supervised models have shown success in tasks like risk stratification and mortality prediction, RL provides a framework for learning sequential decision policies from historical data~\cite{Komorowski2018, kulkarni2022optimizing}. Offline RL, in particular, is well-suited for clinical settings, where real-time exploration is neither feasible nor ethical.

Despite increasing interest in applying RL to ICU settings, most work remains focused on supervised risk scoring or binary classification. Effective RL models must not only learn optimal policies from logged data but also support generalization to out-of-distribution states and produce reliable, interpretable outputs for clinical use.

In this work we present a framework for personalized treatment through RL, synthetic data generation, and language model-based rationale generation. Our main contributions include:
\begin{enumerate}
\item We developed an interpretable offline RL pipeline that combines Advantage-Weighted Regression (AWR) with a simplified attention mechanism. This output is then combined with the ensemble predictions from XGBoost and TabNet to improve learning stability and treatment accuracy while maintaining interpretability.

\item We utilize clustering-driven stratification to group patients by risk using HDBSCAN \cite{McInnes2017}, which allows us to handle the cold-start problem for the patients with limited or no ICU history by assigning them to similar historical trajectories.

\item To address class imbalance and data sparsity in critical interventions (e.g., vasopressors), we augment the dataset with synthetic trajectories generated via a diffusion model and a conditional VAE.
\item We integrate a multi-modal large language model (LLM) into the inference pipeline to generate contextual, patient-specific rationales for selected actions. The model combines current vitals, retrieved clinical knowledge, and RL outputs to support explainable decision-making.
\end{enumerate}

\begin{figure*}[th!]
    \centering
    \includegraphics[width=1\textwidth]{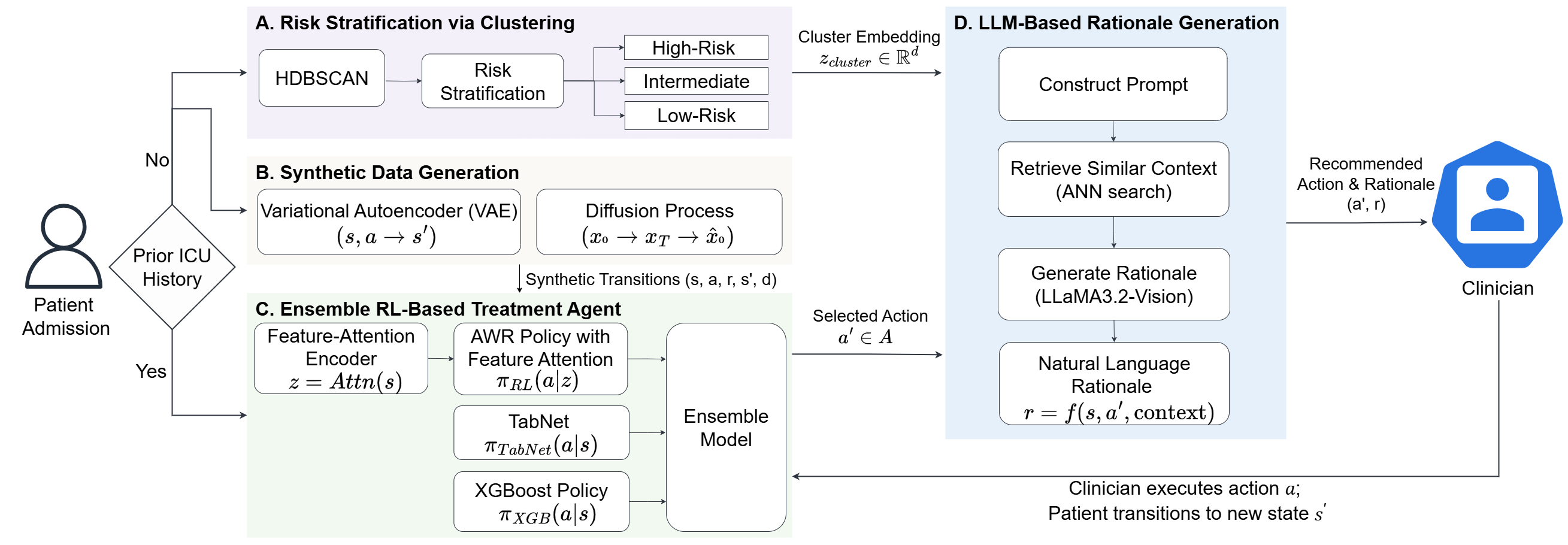}
    \caption{Overview of the interpretable sepsis treatment pipeline. (A) Patients without prior ICU history are stratified into low-, intermediate-, or high-risk groups using clustering. (B) To address data sparsity, synthetic transitions \((s, a, r, s', d)\) are generated using a VAE and a diffusion model, then added to the RL training set. (C) For intermediate-risk or historical patients, a feature-attention encoder produces a latent state \(z = \mathrm{Attn}(s)\), used by an AWR policy \(\pi_{\phi}(a \mid z)\), a Q-network, and a value function. The final recommendation \(a' = \arg\max_{a} [\text{blend}(\pi_{\phi}, \pi_{\text{XGB}})]\) combines outputs from AWR and a clinician-trained XGBoost policy. (D) To enhance interpretability, a local LLM generates a natural-language rationale \(r = f(s, a', \text{context})\) using the patient state, selected action, and retrieved clinical context.}
    \label{fig:main_pipeline}
\end{figure*}

\section{Background and Literature Review}

\subsubsection{Sepsis as a Sequential Decision Problem}
The diagnostic and therapeutic challenges in sepsis, characterized by hidden disease states and incomplete observability of the underlying pathophysiology, allow us to formulate it as a Partially Observable Markov Decision Process (POMDP). The patient’s physiological state evolves in response to administered treatments (e.g., vasopressors, fluids) and latent disease progression. Clinical guidelines emphasize the need for timely intervention, where delays in antibiotic or fluid administration substantially increase risk~\cite{Rhee2019, Evans2021}.

\subsubsection{Challenges in Modeling Sepsis from ICU Data}
Training RL agents in healthcare is constrained by the absence of an online environment and the inability to perform exploration. Offline RL algorithms address this by learning from historical data while correcting for the distributional mismatch between the behavior and the learned policy. However, the quality of the learned policy is closely coupled with state representation and reward design.

ICU datasets such as MIMIC-III \cite{johnson2016mimic} and eICU \cite{Pollard2018} offer time-stamped, high-resolution records of patient vitals, lab tests, interventions, and outcomes. Yet, they reflect evolving clinical standards, e.g., the definition of sepsis changed mid-decade to emphasize organ dysfunction over simple infection markers\cite{Seymour2016}. This requires harmonization techniques, such as clustering-based cohort construction and dimensionality reduction (e.g., UMAP\cite{McInnes2018}, HDBSCAN \cite{McInnes2017}), to ensure valid cross-temporal comparisons and consistent reward attribution.

\subsubsection{Integration of Language Models for Interpretability}
For AI systems to be useful in clinical practice, they need to explain their reasoning in a way that clinicians can trust. Attention mechanisms offer insights by highlighting which features are important, but they often fall short of providing clear justifications. Recently, LLMs, especially those with multi-modal inputs, have made it possible to generate natural-language explanations grounded in both patient data and clinical knowledge~\cite{Guo2023, Kazi2023b}.

\section{Interpretable Sepsis Treatment Methodology}
Our methodology integrates patient risk stratification (Sec.~\ref{sub:risk_stratification}), synthetic data augmentation (Sec.~\ref{sub:data_generation}), offline reinforcement learning (Sec.~\ref{sub:rl_agent}), and LLM-based interpretability (Sec.~\ref{sub:llm_rationale}) to develop a transparent and data-driven sepsis treatment policy. The full pipeline is illustrated in Fig.~\ref{fig:main_pipeline}.

\subsection{Risk Stratification via Clustering}
\label{sub:risk_stratification}
Newly admitted patients often lack sufficient ICU history, making it challenging to apply downstream RL and LLM modules that rely on longitudinal data. To address this, we use unsupervised clustering to assess patient status upon admission, grouping them into risk categories based on initial vitals and lab measurements. Clustering helps identify patients according to their risk stratification. Prior to RL training, patient states are clustered to identify distinct sepsis progression patterns, and recent advances in clustering efficiency, such as the centroid update approach by Borlea et al. \cite{article}, demonstrate significant reductions in computational iterations while maintaining clustering quality, a critical consideration for large-scale patient data preprocessing in clinical settings. We utilize Hierarchical Density-Based Spatial Clustering of Applications with Noise (HDBSCAN) \cite{McInnes2017} and Uniform Manifold Approximation and Projection (UMAP) \cite{McInnes2018} as our core unsupervised clustering and visualization tools. This combination confers several key advantages over conventional clustering algorithms within the medical domain. Specifically, it accommodates variable cluster densities, enables automatic cluster detection, demonstrates robustness to noise and outliers, and is scalable for real-time deployment.

This process broadly categorizes patients into the following:
\begin{enumerate}
    \item \textbf{Low-risk $[0\%, 40\%]$:} Patients with stable vitals and a good recovery trajectory, thus no intervention is needed.
    \item \textbf{Intermediate-risk $(40\%, 75\%]$:} Patients suitable for our RL-based recommendations in conjunction with clinical judgment.
    \item \textbf{High-risk $(75\%, 100\%]$:} Patients requiring immediate clinical intervention.
\end{enumerate}

\begin{figure}[htbp]
    \centering
    \includegraphics[width=0.90\linewidth]{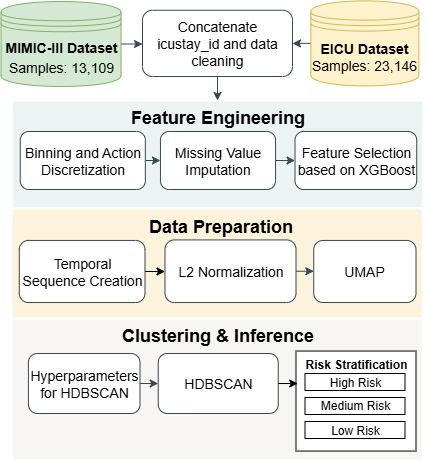}
    \caption{Clustering-based risk stratification pipeline. After preprocessing ICU data, including feature selection, temporal sequence construction, and UMAP Dimensionality reduction, HDBSCAN is used to group patients by similarity. The resulting clusters are validated using risk trends to ensure clinical relevance.}
    \label{fig:clustering_pipeline}
    \vspace{-0.25 cm}
\end{figure}

The pipeline, shown in Fig.~\ref{fig:clustering_pipeline}, performs temporal filtering, L2 normalization, and UMAP\cite{McInnes2018} dimensionality reduction, followed by HDBSCAN-based grouping. This procedure is detailed in Algorithm~\ref{algo:clustering}.

\begin{algorithm}[htbp]
\caption{HDBSCAN-Based Clustering and Validation}
\label{algo:clustering}
\begin{algorithmic}[1]
    \Require Dataset $D$, Feature set $F$, Max sequence length $L=80$, 
    \Ensure Risk categories R $\in$ {Low, Intermediate, High}
    \State Filter $D$ by temporal length; pad sequences to length L
    \State Apply L2 normalization and UMAP for dimensionality reduction; split data into training and test sets: $D = D_{\text{train}} \cup D_{\text{test}}$.
    \State Initialize HDBSCAN: $min\_cluster\_size = m$, $min\_samples = s$, $\epsilon = \epsilon$.
    \State Fit HDBSCAN on $D_{\text{train}}$ to obtain clusters $C_{\text{train}}$ and noise $N_{\text{train}}$.
    \State Predict clusters on $D_{\text{test}}$ using trained model → $C_{\text{test}}, N_{\text{test}}$
    \State Evaluate mortality ($M$) variance: $\text{Var} = \sum_{k=1}^{K} (M_k - M_{\text{overall}})^2 / K$, where $k$ represents the $k^{th}$ cluster.
    \State Compute chi-square statistic: $\chi^2 = \sum_i (O_i - E_i)^2 / E_i$, where $O_i$ and $E_i$ are observed and expected frequencies.
    \State $p-value = P(\chi^2 >= \chi^2_{\text{observed}})$
    \If{$p$-value $< 0.001$}
    \State Accept clusters as statistically significant.
    \EndIf
    \For{each cluster $c$ with mortality rate $m_c$}
    \If{$m_c \leq 0.40$}
        \State assign \textbf{Low Risk}
    \ElsIf{$0.40 < m_c \leq 0.75$}
        \State assign \textbf{Intermediate Risk}
    \Else
        \State assign \textbf{High Risk}
    \EndIf
\EndFor
    \State \Return Risk Categories $R$, Cluster Assignments, Validation Metrics
\end{algorithmic}
\end{algorithm}

This module provides interpretable early risk labels when clinical records are limited by using unsupervised clustering with statistical validation.

\subsection{Synthetic Data Generation}
\label{sub:data_generation}

In the medical domain, datasets are often limited and costly to collect. To address this, we augment MIMIC-III and eICU datasets with realistic synthetic patient trajectories using two complementary generative modeling methods: a diffusion model for continuous-time state transitions, and a Variational Autoencoder (VAE) for modeling discrete transitions conditioned on actions and rewards.

\subsubsection*{\textbf{Diffusion Model}}
We apply a lightweight diffusion process to each normalized observation window \(x_0\) to generate realistic synthetic samples.

\begin{enumerate}
  \item Noise is added incrementally using a schedule \(\{\beta_t\}_{t=1}^{T}\), where \(\beta_t\) gradually increases from near zero to a small positive value.
  
  \item At each step, the data is perturbed with Gaussian noise using the forward kernel (Eq.~\ref{eq:forward_kernel}), gradually increasing the noise level in the sample.
  \begin{equation}
  \label{eq:forward_kernel}
  q(x_t \mid x_{t-1}) = \mathcal{N}\!\bigl(\sqrt{1-\beta_t}\,x_{t-1},\; \beta_t I\bigr)
  \tag{2}
  \end{equation}

  \item A neural network \(\epsilon_\theta(x_t, t)\) is trained to estimate the added noise. The model minimizes the objective represented in Eq.~\ref{eq:nn}:
  \begin{equation}
  \label{eq:nn}
  \mathcal{L}_{\mathrm{diff}} = \mathbb{E}_{x_0, \epsilon, t} \left[ \lVert \epsilon - \epsilon_\theta(x_t, t) \rVert^2 \right]
  \tag{3}
  \end{equation}

  \item To generate a new sample, the process starts from pure noise \(x_T \sim \mathcal{N}(0, I)\) and iteratively applies the denoiser from \(t = T\) down to \(t = 0\), yielding a reconstructed window \(\hat{x}_0\).

  \item Finally, the output \(\hat{x}_0\) is unscaled, clipped to remove implausible values, and added to the synthetic dataset used for training the RL agent.
\end{enumerate}

\subsubsection*{\textbf{Variational Autoencoder (VAE)}}

Discrete, action-conditioned transitions are modeled using a conditional VAE. The VAE captures complex dependencies between the current state \(s\), action \(a\), reward \(r\), terminal indicator \(d\), and the next state \(s'\).

The encoder network \(f_{\phi}\) maps the input pair \((s, a)\) to a latent distribution \(q_{\phi}(z \mid s, a) = \mathcal{N}(\mu, \sigma^2)\), where \((\mu, \log\sigma^2) = f_{\phi}(s, a)\). Latent samples are drawn using the reparameterization trick: \(z = \mu + \sigma \odot \epsilon\), with \(\epsilon \sim \mathcal{N}(0, I)\). The decoder \(g_{\psi}\) then reconstructs the next state \(\hat{s}'\) conditioned on the latent variable along with the action, reward, and terminal indicator: \(\hat{s}' = g_{\psi}(z, a, r, d)\). Training minimizes a loss function that combines reconstruction error and KL divergence:
\[
\mathcal{L}_{\mathrm{VAE}}(\phi, \psi) = \mathbb{E}_{q_{\phi}(z|s,a)}[\| s' - \hat{s}' \|^2] + \beta\, D_{\mathrm{KL}}(q_{\phi}(z|s,a)\,||\,p(z)),
\]
where \(p(z) = \mathcal{N}(0, I)\) is the standard Gaussian prior and \(\beta\) controls the regularization strength. Prior to training, all states are normalized, and discrete actions are encoded as one-hot vectors to match the input requirements of the encoder.

We generate synthetic transitions to address a class-imbalance problem as follows:

\begin{enumerate}
    \item Sample latent variables \(z \sim \mathcal{N}(0, I)\) from the prior distribution.
    \item Select action \(a\), reward \(r\), and terminal flag \(d\) from empirical distributions.
    \item Decode using \(g_{\psi}(z, a, r, d)\) to obtain a synthetic next state \(\hat{s}'\).
    \item Apply post-processing to clip outliers and filter implausible transitions.
\end{enumerate}

This combined approach of diffusion modeling and VAE helps to generate realistic synthetic trajectories for underrepresented interventions like vasopressors and fluids. 

\begin{figure}[htbp]
    \centering \includegraphics[width=1\linewidth]{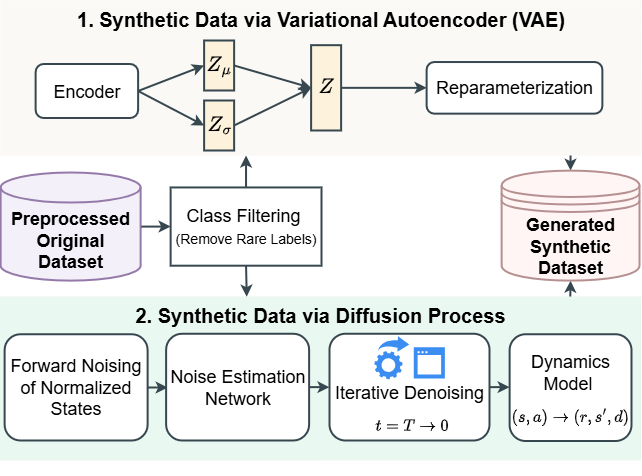}
    \caption{Minority-class augmentation pipeline. Cleaned MIMIC-III + eICU records are filtered to classes 1 \& 2, then two parallel generators create new samples: (1) a VAE that decodes latent draws, and (2) a conditional diffusion process that iteratively denoises scaled states with a time-stepped loop.}
    \label{fig:synthetic_data_pipeline}
    \vspace{-0.25 cm}
\end{figure}

\subsection{Ensemble RL-Based Treatment Agent}
\label{sub:rl_agent}

We apply offline RL to optimize sepsis treatment policies using patient data from the MIMIC-III and e-ICU database. Due to data sparsity, especially in vasopressors and fluid intervention categories, we supplement training with synthetic trajectories generated via VAE and diffusion models (Sec.~\ref{sub:data_generation}). 

\begin{figure*}[htbp]
  \centering
  \includegraphics[width=\textwidth]{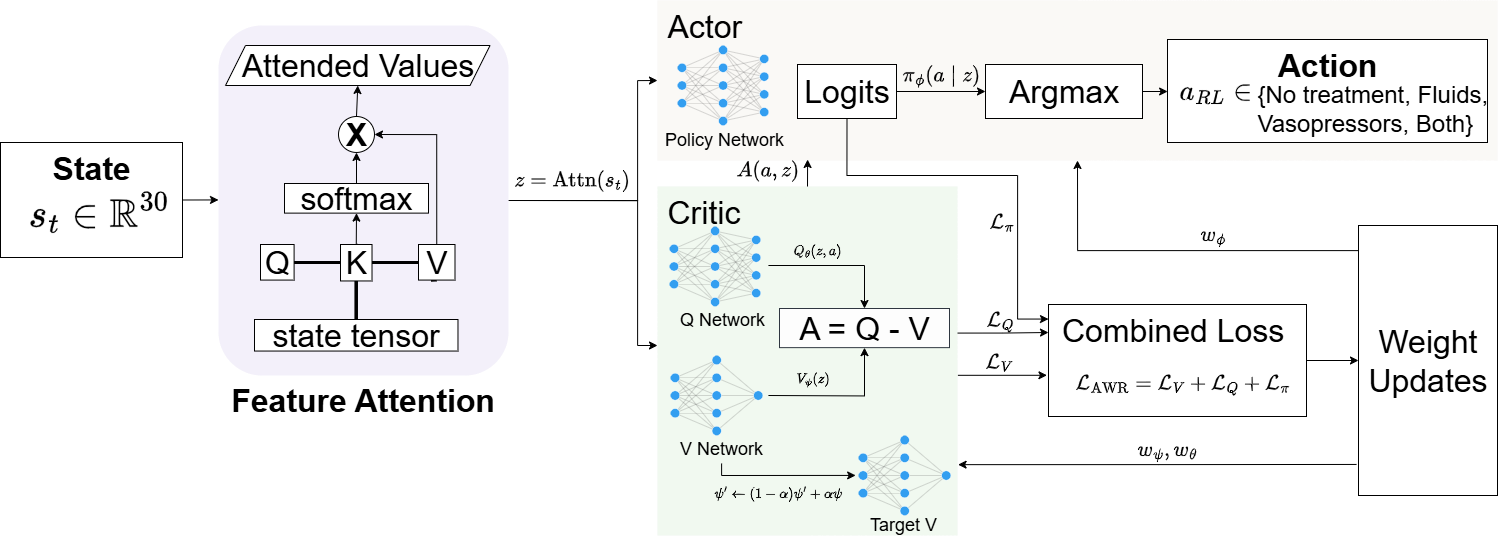}
  \caption{AWR with a custom feature attention pipeline. The $30‑$D state goes through a custom attention block and a shared encoder. Policy ($\pi$), Q, and V heads are trained together. We weight the policy loss using the advantage value and use a target-value network to smooth the learning updates.}
  \label{fig:rl-new}
\end{figure*}

\subsubsection*{\textbf{Feature Preparation}}

The following steps are used to prepare the dataset using Google BigQuery:

\begin{enumerate}
  \item Extract adult ICU stays with a sepsis diagnosis from MIMIC-III and eICU.
  \item Split each patient’s time series of vital signs and lab values into fixed-length windows of \(4\) hours.
  \item Fill any missing measurement by carrying the last observation forward; if still missing, use the median value for that feature.
  \item Scale each feature \(x_j\) using its training-set mean \(\mu_j\) and standard deviation \(\sigma_j\) as $\tilde{x}_j \;=\; \frac{x_j - \mu_j}{\sigma_j}$.
\end{enumerate}

\subsubsection*{\textbf{Feature Importance Analysis via XGBoost}}
To validate our data preprocessing and imputation strategy, we conducted a feature importance analysis using an XGBoost classifier. Specifically, we augmented the state features with a synthetic ``noise" feature, randomly generated from a uniform distribution. The XGBoost model is trained to predict the discrete action classes (four actions in our RL environment) based on the state features.

\subsubsection*{\textbf{RL Environment}}

\paragraph{Observation Space}
Each patient state \(s_t \in \mathbb{R}^{d}\) captures a snapshot of the patient’s condition at time step \(t\), encoded as a vector of dimension \(d = 30\). The feature set includes vital signs (e.g., heart rate, mean arterial pressure [MAP], oxygen saturation [SpO\(_2\)]), laboratory values (e.g., lactate, creatinine, white blood cell count), and treatment indicators (e.g., prior administration of vasopressors or fluids). All features are standardized using z-score normalization and missing values are imputed using domain-aware techniques.

\paragraph{Action Space}

The agent selects from a discrete action space with four treatment choices, representing common sepsis treatment strategies:
{\small
\[
\mathcal{A} = \{ \text{No treatment},\ \text{Fluids},\ \text{Vasopressors},\ \text{Combined treatments} \}
\]
}

\paragraph{Reward Function} \label{sec:reward}
The reward function is designed to align with key treatment goals: stabilizing vital signs and reducing short-term mortality. At each step, the agent receives a composite reward:

\begin{equation}
\begin{aligned}
r_t &= -\, I\{\text{mortality within 48 h}\} + 0.3\, I\{\text{MAP}_t > 65\} \\
    &\quad + 0.3\, I\{\text{SpO}_2{}_t > 94\} + 0.2\, I\{\text{lactate}_t < 2\}
\end{aligned}
\tag{4}
\end{equation}

\noindent where $I$ is the scaling factor~\cite{Komorowski2018}. The reward ranges from \(-1\) (next \(48-\)h mortality) to \(+0.8\) when all three vitals are in safe ranges, including hemodynamic stability (MAP), adequate oxygenation (SpO\(_2\)), and reduced metabolic stress (lactate).  This reward function encourages the agent to prefer treatments that stabilize the patient in the short-term, while staying closely aligned with clinically meaningful goals.

\subsubsection*{\textbf{AWR with Feature Attention}}
We apply Advantage-Weighted Regression (AWR)~\cite{peng2019advantage} with a lightweight attention-based encoder. The attention module transforms raw state vectors \(s\) into latent embeddings \(z = \mathrm{Attn}(s)\) to capture the most important features for the RL agent.

\begin{enumerate}
  \item We compute the next attended state \(z'=\mathrm{Attn}(s')\) and form the one‐step value target:
    \[
      y_{V}
      = r \;+\; \gamma\,(1 - d)\;V_{\psi'}(z')\,,
      \quad
      \delta = y_{V} - V_{\psi}(z)\,
      \tag{5}
    \]
  \item We train the value head with an \emph{expectile regression}:
    \[
      \mathcal{L}_{V}
      = \mathbb{E}\bigl[w(\delta)\,\delta^{2}\bigr],
      \quad
      w(\delta)=
      \begin{cases}
        \tau, & \delta > 0,\\
        1-\tau, & \delta < 0
      \end{cases}
      \tag{6}
    \]
    which pushes \(V_{\psi}(z)\) toward the bootstrap target \(y_{V}\) while weighting over‐ and under‐estimates differently.
  \item We train the Q‐head to match the same target:
    \[
      \mathcal{L}_{Q}
      = \mathbb{E}\Bigl[\bigl(Q_{\theta}(z,a) - y_{V}\bigr)^{2}\Bigr]
      \tag{7}
    \]
  \item We form the \emph{advantage} \(A = Q_{\theta}(z,a) - V_{\psi}(z)\), convert it into a weight
    \(\displaystyle w_A = \exp\bigl(A / \beta\bigr)\)
    and train the policy by:
    \[
      \mathcal{L}_{\pi}
      = -\,\mathbb{E}\bigl[w_A\,\log \pi_{\phi}(a\mid z)\bigr]
      \tag{8}
    \]
\end{enumerate}

The full training procedure is outlined in Algorithm~\ref{alg:awr_feature_attention}, which details how each transition \((s, a, r, s', d)\) is used to update the policy, value, and Q-networks.

\begin{algorithm}[htbp]
\caption{Advantage-Weighted Regression (AWR) with Feature Attention}
\label{alg:awr_feature_attention}
\begin{algorithmic}[1]
\Require Batch of transitions $(s, a, r, s', d)$, discount factor $\gamma$, expectile $\tau$, temperature $\beta$, soft update rate $\alpha$
\Ensure Updated network parameters $(\psi, \theta, \phi)$ for value, Q, and policy networks
\ForAll{transition $(s, a, r, s', d)$ in batch}
    \State Encode current and next states with attention: $z \gets \mathrm{Attn}(s)$, $z' \gets \mathrm{Attn}(s')$
    
    \State Compute target value:$y_V \gets r + \gamma \cdot (1 - d) \cdot V_{\psi'}(z')$
    \State Compute TD error: $\delta \gets y_V - V_{\psi}(z)$    
    \State Compute value loss: $\mathcal{L}_V \gets \mathbb{E}[w(\delta) \cdot \delta^2]$ where
    \[
      w(\delta) = 
      \begin{cases}
        \tau, & \delta > 0 \\
        1 - \tau, & \delta < 0
      \end{cases}
    \]
    \State Compute Q-value estimate: $Q_{\theta}(z, a)$
    \State Compute Q loss:$\mathcal{L}_Q \gets \mathbb{E}[(Q_{\theta}(z, a) - y_V)^2]$
    \State Compute advantage: $A \gets Q_{\theta}(z, a) - V_{\psi}(z)$
    \State Compute weight: $w_A \gets \exp(A / \beta) / \mathbb{E}[\exp(A / \beta)]$
    \State Compute policy loss: $\mathcal{L}_\pi \gets -\mathbb{E}[w_A \cdot \log \pi_{\phi}(a \mid z)]$
    \State Compute total loss: $\mathcal{L} \gets \mathcal{L}_V + \mathcal{L}_Q + \mathcal{L}_\pi$
    \State Backpropagate and update all networks using $\mathcal{L}_{\text{AWR}}$
    \State Soft update target: $\psi' \gets (1 - \alpha) \cdot \psi' + \alpha \cdot \psi$
\EndFor
\State \Return Updated parameters $(\psi, \theta, \phi)$
\end{algorithmic}
\end{algorithm}

\subsubsection*{\textbf{Ensemble-Based Treatment Policy}}
Our final treatment decision is made by an ensemble of three models: an AWR-based RL agent, TabNet, and XGBoost. Because these decisions involve sensitive interventions like vasopressors and fluids, we implement a conservative, safety-oriented rule: a treatment is recommended if any model in the ensemble predicts a probability for it that exceeds a set threshold ($\omega$).

Let $p_{\text{fluid}}$ and $p_{\text{vaso}}$ be the maximum predicted probabilities (across TabNet and XGBoost) for fluid and vasopressor treatments, and let $a_{\text{RL}}$ be the action suggested by the RL agent. The final action $a^*$ is computed as:

\[
a^* = 
\begin{cases}
\text{fluid}, & \text{if } p_{\text{fluid}} > \omega \text{ and } p_{\text{fluid}} > p_{\text{vaso}} \\
\text{vasopressor}, & \text{if } p_{\text{vaso}} > \omega \text{ and } p_{\text{vaso}} > p_{\text{fluid}} \\
a_{\text{RL}}, & \text{otherwise}
\end{cases}
\tag{9}
\]

The recommendations from the two tabular models, TabNet and XGBoost, are evaluated first. If either model's confidence for a treatment exceeds a predefined threshold ($\omega$), that action is taken, overriding the RL agent. Otherwise, the final decision is delegated to the learned RL policy. This hybrid approach ensures a safety-aware framework that balances data-driven decisions with clinical policies.

\subsubsection*{\textbf{TabNet}}

It is a deep learning model specifically designed for structured tabular data. It uses sequential attention to focus on the most relevant features at each decision step. This makes it useful for interpretability and handling feature sparsity.

Given an input state vector $x \in \mathbb{R}^d$, TabNet applies a series of attention-based decision steps to produce feature masks and predictions. At each step $t$, a mask $M^{(t)}$ is generated to select a subset of features: $M^{(t)}~=~\text{Sparsemax}(P^{(t)})$, where $P^{(t)}~=~\text{Attn}(x^{(t)})$.

The selected features are then passed through shared decision layers to refine the output. The final treatment prediction is obtained after aggregating the outputs across all steps.

\subsubsection*{\textbf{XGBoost}}
It is a gradient-boosted decision tree model that is well-suited for our data. We use it as a secondary model to estimate the probability of recommending each treatment action based on patient state features.

Given a state input $x \in \mathbb{R}^d$, XGBoost builds an ensemble of decision trees to learn the probability distribution over discrete treatment classes:
\[
\hat{y} = \sum_{m=1}^M f_m(x), \quad f_m \in \mathcal{F}
\tag{10}
\]
where $\mathcal{F}$ is the space of regression trees, and each $f_m$ is a tree added at iteration $m$. The final prediction $\hat{y}$ is interpreted as a softmax probability across the treatment classes.

\subsection{LLM-Based Rationale Generation}
\label{sub:llm_rationale}

To improve the interpretability of clinical decision-making, we integrate a multi-modal large language model into the inference pipeline. The goal is to generate natural-language rationales that explain why a particular treatment is selected; this way, we aim to increase transparency and clinicians' trust.

\begin{figure}[htbp]
  \centering
  \includegraphics[width=1\linewidth]{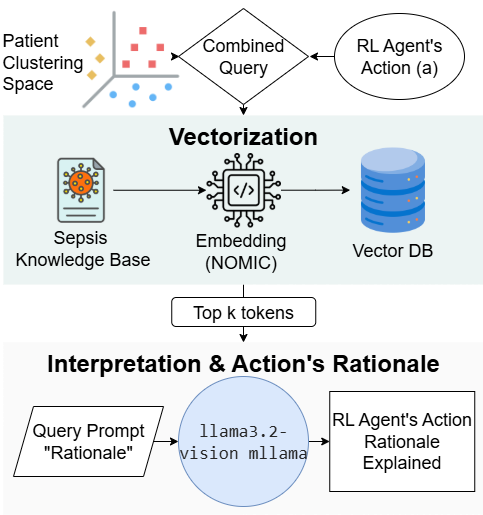}
  \caption{Overview of the LLM-based rationale generation process. A query, combining the patient's cluster state and the RL agent's action, is used to retrieve relevant context from a vectorized sepsis knowledge base. The retrieved information is then passed to an LLM to generate a final, knowledge-grounded explanation for the action.}
  \label{fig:architecture}
  \vspace{-0.25 cm}
\end{figure}

We construct a natural-language prompt (Fig.~\ref{fig:architecture}) from the patient's clinical context as follows:

\begin{enumerate}
    \item The patient's current state ($s$) is mapped to a point in the learned patient clustering space. We merge this state with the selected action $a \in A$ from the ensemble-based RL policy.
    \item The combined query is then used to search a Vector DB via an Approximate Nearest Neighbor (ANN) search. The search retrieves the top-k most relevant tokens from an expert-curated sepsis knowledge base that has been pre-vectorized using a NOMIC encoder \cite{Rahman2023}.
    \item The retrieved knowledge tokens are inserted into a prompt template. This prompt is passed to the LLM, which generates a clear, natural-language rationale explaining the clinical reasoning for the recommended action ($a$) in the context of the patient's state ($s$).
\end{enumerate}

\section{Evaluation Results \& Analysis}
\subsection{Datasets}

Large-scale, de-identified critical care databases such as MIMIC-III and the eICU Collaborative Research Database have become foundational resources for sepsis research and clinical modeling. The Medical Information Mart for Intensive Care III (MIMIC-III)~\cite{johnson2016mimic} contains detailed records from over $53\,000$ ICU admissions at a single tertiary care hospital in Boston, collected between 2001 and 2012. The patient population primarily consists of adults, with a median age of 66 and an overall in-hospital mortality rate of approximately $11\%$. In contrast, the eICU database includes more than $200\,000$ ICU stays across 208 U.S. hospitals from 2014 to 2015.
Both datasets provide high-resolution, time-stamped clinical information, including vital signs, lab results, interventions, diagnostic codes, and outcomes.

\subsection{Latent Space Preparation}

In order to identify clinically meaningful patient subgroups, we performed unsupervised clustering over a merged dataset combining MIMIC-III~\cite{johnson2016mimic} and eICU~\cite{Pollard2018}. After aligning on \texttt{icustay\_id}, the merged dataset contained $874\,108$ time-stamped records across $27\,799$ ICU stays.

\paragraph{Temporal Filtering and Preprocessing}
We restricted our analysis to patients with between $1$ and $80$ time points, which preserved approximately $75\%$ of the original data. This range was empirically chosen to exclude extremely long hospital stays (e.g., several months), which are atypical and prone to introducing temporal noise. The resulting dataset retained all $27\,799$ unique patients. Each patient's temporal trajectory was converted into a structured sequence with zero-padding to handle varying sequence lengths and create uniform dimensions. After padding, the final feature matrix size was $25\,605 \times 320$, encompassing physiological variables and demographic attributes such as `\textit{spo2}', `\textit{platelets}', and `\textit{hours since ICU admission}'.

\paragraph{Dimensionality Reduction and Clustering}
To enable efficient clustering in a lower-dimensional latent space, we applied Uniform Manifold Approximation and Projection (UMAP)\cite{McInnes2018}, as it better preserves the global structure of the data~\cite{Cavalcant2024L1NG, 10903282}, and this in turn helps with visualizing the clusters and better understanding the patterns it has. We then applied HDBSCAN~\cite{McInnes2017}, a density-based clustering algorithm robust to noise and varying cluster densities. After hyperparameter tuning, we determined optimal clustering parameters: minimum cluster size = $30$; minimum samples = $30$; epsilon~=~$0.01$.

This yielded $124$ distinct clusters during training. For interpretability and downstream integration, we grouped them into three clinical risk categories: low, intermediate, and high~\cite{Ahmad2023HC3FL}. Noisy data comprised $5.3\%$ of training data and $11.3\%$ of test data, which is acceptable given the complexity and variability inherent in clinical data. Our approach is similar to methods used in behavioral health monitoring \cite{Radhakrishnan2024Behavioral} and neuro-dynamics clustering~\cite{Khoshkhahtinat2023Epileptic}.

Overall clusters exhibited notable distinction in terms of mortality risk within 48 hours (Table~\ref{tab:cluster_mortality}).

\begin{table}[htbp]
\small
\centering
\caption{Risk Stratification for Clusters}
\begin{tabular}{>{\centering\arraybackslash}m{2.4cm}|
                >{\centering\arraybackslash}m{1.3cm}|
                >{\centering\arraybackslash}m{1.2cm}|
                >{\centering\arraybackslash}m{2.3cm}}
\toprule
\textbf{Cluster} & \textbf{Mortality (\%)} & \textbf{Patients} & \textbf{Risk Category} \\
\midrule
1, 86--123 & 100.0 & 31--211 & High Risk \\
\midrule
0, 2, 6, 43, 54, 59--62, 65 & 32.6--62.4 & 62--220 & Intermediate Risk \\
\midrule
7, 12, 13, 15, 19, 24, 27, 28, 40, 42, 64, 75, 79, 80, 83, 97, 108 & 0.0--4.2 & 31--836 & Low Risk \\
\bottomrule
\end{tabular}
\label{tab:cluster_mortality}
\vspace{-0.25 cm}
\end{table}

\subsection{Reinforcement Learning Experiments}

\subsubsection*{\textbf{Feature Importance Analysis}}

Before evaluating RL models, we performed feature importance analysis using XGBoost to validate our preprocessing and data imputation strategy. Features were ranked using the average gain metric across decision trees. Clinically meaningful variables such as SpO\textsubscript{2}, platelets, and MAP dominated the top rankings (Table~\ref{tab:feature_importance}). In contrast, a synthetic random noise feature received the lowest importance score, confirming that the model appropriately distinguishes signal from noise.

\begin{table}[htbp]
\centering
\caption{Top Features by XGBoost}
\label{tab:feature_importance}
\begin{tabular}{lr}
\toprule
\textbf{Feature} & \textbf{Gain Score} \\
\midrule
SpO\textsubscript{2} & 910.93 \\
Platelets & 732.08 \\
MAP & 239.74 \\
Hours Since ICU Admission & 179.62 \\
GCS Total & 138.09 \\
Ethnicity & 116.90 \\
Systolic BP & 104.86 \\
Bicarbonate & 70.51 \\
Random Noise & 0.93 \\
\bottomrule
\end{tabular}
\vspace{-0.10 cm}
\end{table}

\subsubsection*{\textbf{AWR with Attention Mechanism}}
\begin{figure}[htbp]
    \centering
    \includegraphics[width=1\linewidth]{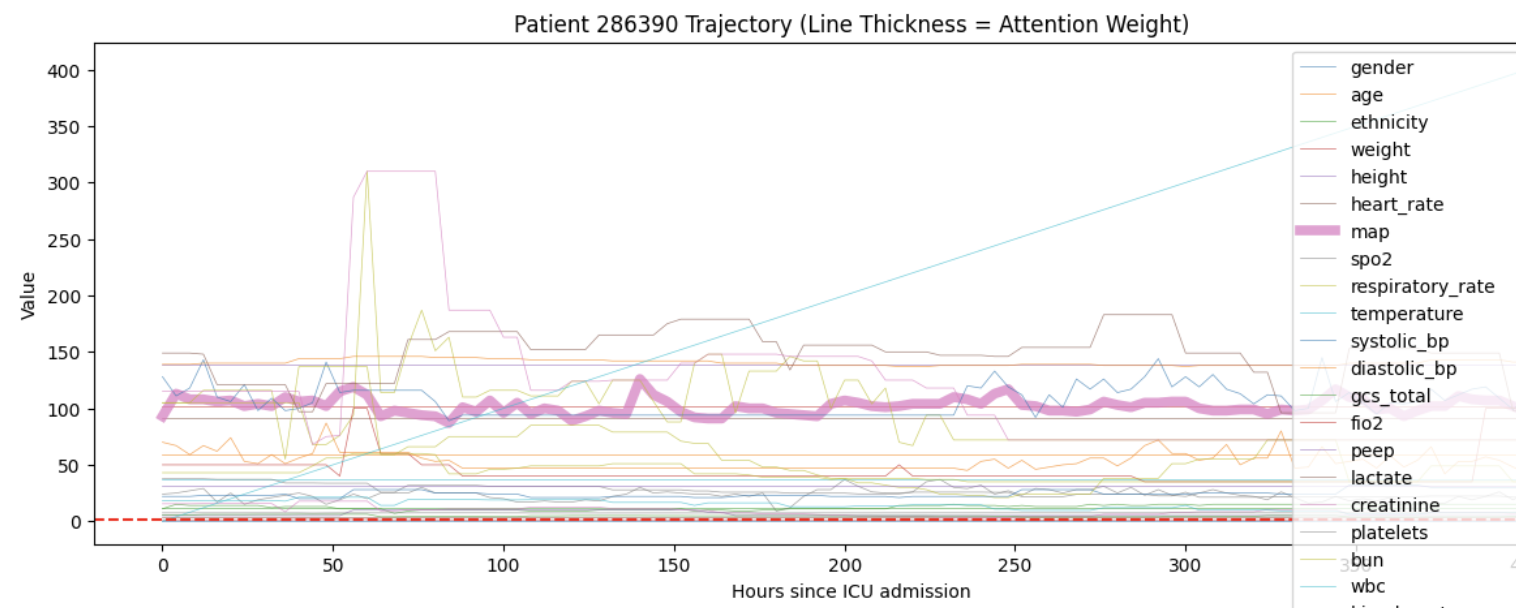}
    \caption{Attention plot for patient trajectory. Each line is a feature over hours since ICU admission, and line thickness is proportional to the attention weight. The model places the greatest emphasis on MAP (bold magenta) with secondary, thinner weights on other vitals/labs.}
    \label{fig:attention-plot}
    \vspace{-0.25 cm}
\end{figure}

Our core RL method is Advantage-Weighted Regression (AWR), augmented with a lightweight attention mechanism. The attention module highlights salient features in each patient state ($s$), producing an attended representation ($z$).
The attention mechanism allows the agent to dynamically prioritize key variables, such as MAP or lactate trends, when deciding treatment strategies. Fig.~\ref{fig:attention-plot} visualizes one such trajectory, showing high attention on MAP and fluid-related features during hypotensive episodes.

To evaluate model performance, we compared several variants of our pipeline:

\begin{itemize}
\item \textbf{BCQ (Behavior Cloning + Q-learning)}: A baseline offline RL method.
\item \textbf{BCQ + Attention}: Adds the attention mechanism to improve state representation.
\item \textbf{AWR + Attention}: Our proposed interpretable offline RL agent.
\item \textbf{Ensemble}: Combines AWR, XGBoost, and TabNet to improve robustness.
\end{itemize}

\noindent\textit{Overall Performance:} AWR + Attention outperforms BCQ in both accuracy and average reward (Table~\ref{tab:combined_ablation}). The ensemble model achieves the best overall accuracy (83\%) by integrating tabular and RL models.

\begin{table}[htbp]
\centering
\small
\caption{Combined Ablation Study}
\label{tab:combined_ablation}
\begin{tabular}{>{\arraybackslash}m{4.1cm}
                >{\centering\arraybackslash}m{1.5cm}
                >{\centering\arraybackslash}m{1.5cm}}
\toprule
\textbf{Model} & \textbf{Accuracy (\%)} & \textbf{Average Reward} \\
\midrule
BCQ & 60 & -0.60 \\
BCQ + Attention & 74 & -0.47 \\
AWR + Attention & 80 & -0.33 \\
Ensemble (XGBoost + AWR + Attention + TabNet) & \textbf{83} & N/A \\
\bottomrule
\end{tabular}
\vspace{-0.10 cm}
\end{table}

\noindent\textit{Per-Class Treatment Performance:}
Table~\ref{tab:precision_recall_study} represents class-wise precision and recall for the four treatment actions (A0: no treatment, A1: fluids, A2: vasopressors, A3: combined). The ensemble model consistently yields the best performance, especially for minority actions (A1, A2), which are underrepresented in the dataset but clinically critical.

\begin{table}[htbp]
\centering
\small
\caption{Precision and Recall for Treatment Actions}
\label{tab:precision_recall_study}
\setlength{\tabcolsep}{3 pt}
\renewcommand{\arraystretch}{1.2}
  \resizebox{\columnwidth}{!}{
\begin{tabular}{lcccccccc}
\toprule
\multirow{2}{*}{\textbf{Model}} & \multicolumn{2}{c}{\textbf{A0}} & \multicolumn{2}{c}{\textbf{A1}} & \multicolumn{2}{c}{\textbf{A2}} & \multicolumn{2}{c}{\textbf{A3}} \\
\cmidrule(lr){2-3} \cmidrule(lr){4-5} \cmidrule(lr){6-7} \cmidrule(lr){8-9}
& \textbf{P} & \textbf{R} & \textbf{P} & \textbf{R} & \textbf{P} & \textbf{R} & \textbf{P} & \textbf{R} \\
\midrule
BCQ & 0.72 & 0.75 & 0.01 & 0.03 & 0.30 & 0.25 & 0.50 & 0.65 \\
BCQ + Attention & 0.82 & 0.84 & 0.03 & 0.10 & 0.45 & 0.40 & 0.60 & 0.72 \\
AWR + Attention & 0.85 & 0.89 & 0.05 & 0.15 & 0.55 & 0.45 & 0.64 & 0.78 \\
XGBoost & 0.84 & 0.86 & 0.38 & 0.69 & 0.55 & 0.48 & 0.60 & 0.74 \\
TabNet & 0.85 & 0.81 & 0.48 & 0.54 & 0.62 & 0.56 & 0.50 & 0.80 \\
Ensemble & \textbf{0.93} & \textbf{0.92} & \textbf{0.50} & \textbf{0.60} & \textbf{0.65} & \textbf{0.60} & \textbf{0.81} & \textbf{0.70} \\
\bottomrule
\end{tabular}}
\vspace{-0.25 cm}
\end{table}

Our results show that integrating an attention mechanism with offline RL improves treatment accuracy and action balance, particularly for underrepresented interventions.

\subsubsection*{\textbf{LLM-Based Rationale Generation}}
Clinical applications demand secure and private deployments; thus, our primary criterion was the ability to run the model locally. We evaluated several LLMs with a focus on offline installation and domain adaptability and selected the multi-modal LLaMA3.2-Vision model for its strong performance and compatibility with offline, healthcare-specific use cases.

The full prompt is passed to the model with top-K sampling set to $100$, repeat penalty of $1.1$, and temperature of $4.7$.

The model produced clinically sound rationales. Two sample outputs include:

\begin{quote}
  \textit{``Vasopressor therapy was initiated due to persistent hypotension (MAP \( < 65 \) mmHg) and elevated lactate, suggesting ongoing hypoperfusion.''}
\end{quote}
or, in a different scenario:
\begin{quote}
  \textit{``No immediate action was taken as vital signs are stable and lactate levels are normal.''}
\end{quote}

These responses showcase the model's ability to generate clinically coherent treatment justifications.

\section{Conclusion}
In this work, we tackled the complex challenge of optimizing sepsis treatment by combining offline reinforcement learning with attention mechanisms, risk-based patient clustering, and natural-language rationale generation using large language models. Overall our system achieved $83~\%$ treatment accuracy.

To address data imbalance and limited coverage of rare but critical treatments, we used synthetic data generated by diffusion models and VAEs. Our clustering-based risk stratification also helps generalize recommendations to new patients, even with minimal history. The system explains its actions in plain language based on each patient’s condition. This helps build clinician trust and supports transparency for auditing. 

\bibliographystyle{IEEEtran}
\bibliography{references}

\end{document}